\documentclass[10pt,twocolumn,letterpaper]{article}

\usepackage{wacv}
\usepackage{times}
\usepackage{epsfig}
\usepackage{graphicx}
\usepackage{amsmath}
\usepackage{amssymb}
\usepackage{tabularx}
\usepackage{multirow}
\usepackage{authblk}
\usepackage{algorithmic}
\usepackage{algorithm}
\wacvfinalcopy % *** Uncomment this line for the final submission
\newcommand*\samethanks[1][\value{footnote}]{\footnotemark[#1]}
\ifwacvfinal\fi
\begin{document}

%%%%%%%%% TITLE
\title{Detecting Face2Face Facial Reenactment in Videos}

% Authors at the same institution
%\author{First Author \hspace{2cm} Second Author \\
% Authors at different institutions
\author[1]{\rm Prabhat Kumar\thanks{This study has been performed when the authors were at IIIT-Delhi.}}
\author[2]{\rm Mayank Vatsa\samethanks}
\author[2]{\rm Richa Singh\samethanks}
\affil[1]{Indian Institute of Science Bangalore}
\affil[2]{Indian Institute of Technology Jodhpur}
\affil[1]{\tt\small prabhatkumar@iisc.ac.in}
\affil[2]{\tt\small {\{mvatsa,richa\}@iitj.ac.in}}

% \author{Prabhat Kumar \\
% IISc Bangalore\\
% {\tt\small prabhatkumar@iisc.ac.in}
% \and
% Mayank Vatsa \hspace{1cm} Richa Singh \\
% IIT Jodhpur\\
% {\tt\small {mvatsa,richa}@iitj.ac.in}
% }

% \author[1]{\rm Prabhat Kumar\thanks{This study has been performed when the authors were at IIIT-Delhi}
% }

% \author[2]{Mayank Vatsa\footnotemark[1]}

% \author[2]{Richa Singh\footnotemark[1]}

\maketitle
\ifwacvfinal\thispagestyle{empty}\fi

%%%%%%%%% ABSTRACT
\begin{abstract}
Visual content has become the primary source of information, as evident in the billions of images and videos, shared and uploaded on the Internet every single day. This has led to an increase in alterations in images and videos to make them more informative and eye-catching for the viewers worldwide. Some of these alterations are simple, like copy-move, and are easily detectable, while other sophisticated alterations like reenactment based DeepFakes are hard to detect.
Reenactment alterations allow the source to change the target expressions and create photo-realistic images and videos. While the technology can be potentially used for several applications, the malicious usage of automatic reenactment has a very large social implication. It is therefore important to develop detection techniques to distinguish real images and videos with the altered ones. This research proposes a learning-based algorithm for detecting reenactment based alterations. The proposed algorithm uses a multi-stream network that learns regional artifacts and provides a robust performance at various compression levels. We also propose a loss function for the balanced learning of the streams for the proposed network. The performance is evaluated on the publicly available FaceForensics dataset. The results show state-of-the-art classification accuracy of 99.96\%, 99.10\%, and 91.20\% for no, easy, and hard compression factors, respectively.
\end{abstract}

%%%%%%%%% BODY TEXT
\section{Introduction}

Approximately 95 million photos and videos are uploaded daily on Instagram \cite{instagramstats}. YouTube receives 300 hours of video uploads every minute, with about 5 billion views every single day \cite{youtubestats}. These visual contents, on one hand, act as a medium to interact with individuals, share opinions and thoughts, and reach out to the public. On the other hand, it also serves as a source of information and entertainment. This two-way exchange makes videos and images an effective form of communication between the creators and the viewers. These images and videos are not always posted in the original form but, more often than not, are altered to make them more eye-pleasing for the viewer \cite{aparna-tifs}. It is primarily done by the use of filters available to the creator or by editing software such as Photoshop. These include alterations such as splicing and copy-move. However, some recent opes are more advanced and sophisticated, and lie under the category of ``DeepFakes". Deepfakes, as the name suggests, are often the result of video synthesis commonly done by the use of deep learning networks. Deepfakes include alterations of two kinds - identity swap and reenactment.

% \footnotetext[1]{}

\begin{figure}[tp]
\begin{center}
\includegraphics[width=8cm]{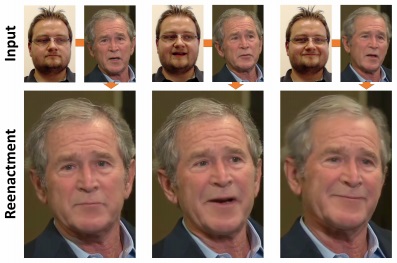}
\end{center}
   \caption{Effect of Reenactment by Face2Face \cite{thies2016face2face}, the source actor (left), the target actor(right), reenactment of the target actor based upon source actor (bottom).}
\label{fig:face2face}
\end{figure}

Reenactment is defined as the \textit{acting out of a past event}; in other words, performing a past event or, with modifications as required.  Facial reenactment refers to the modifications brought to the target actions in the form of change of movement of the head, lips, and facial expression. The techniques allowing for reenactment have been devised with the intent of
improving the experience, specifically in the case of movies with the dubbing of target actors \cite{garrido2015vdub,suwajanakorn2017synthesizing} and teleconferencing \cite{thies2016face2face,thies2018headon}. However, the malicious use of such techniques cannot be ruled out. Specifically, reenactment techniques are capable of synthesizing photo-realistic videos and images that are hard to detect with the human eye or even with  existing forgery detection techniques. Data compression also adds to the challenge of the detection task as often, the media in circulation are highly compressed and offer little knowledge of being altered.

Despite the increased awareness about fake news, videos, and images still remain one of the most trustable sources of information. Reenacted video, as seen in Figure \ref{fig:face2face}, can be used to portray an individual saying things that he/she has not said in the real life. Such videos circulated to a billion uninformed audience via the Internet can lead to chaos and confusion at a large scale. With very limited prior work done in detection, there is an urgent need for developing techniques that can be used for the detection of such alterations.

This paper addresses the problem of detecting reenactment in videos. Our contributions are two-fold; (i) we propose a multistream deep learning network based on the extraction of localized features for detection of reenacted frames by Face2Face reenactment approach \cite{thies2016face2face} in videos, and (ii) we propose a loss function for balanced training of streams in the proposed network. The paper has been organized as follows: Section \ref{related_work} expands upon the generation and detection of reenactment video through subsections \ref{generation_tech} and \ref{detection_tech}, respectively. In section \ref{method} we explain the pipeline, including the deep learning architecture for successful detection of Face2Face reenactment \cite{thies2016face2face}. In section \ref{dataset}, we provide the description of the dataset used for experiments, and the results of the experiments are discussed in section \ref{results}.

\section{Related Work}
\label{related_work}

Attacking visual content using a deep learning approach and their defense is an important area of research \cite{agarwal2018image,goel2018smartbox,Goswami2019,goswami2018unravelling,majumdar2019evading}. The face reenactment literature can be categorized into two broad categories: (i) the generation techniques implying the methods that pave the way for reenactment manipulation on videos or in some cases images and (ii) detection techniques aimed at detecting such forgeries in videos and images.

\begin{figure}[tb]
\begin{center}
\includegraphics[width=8cm]{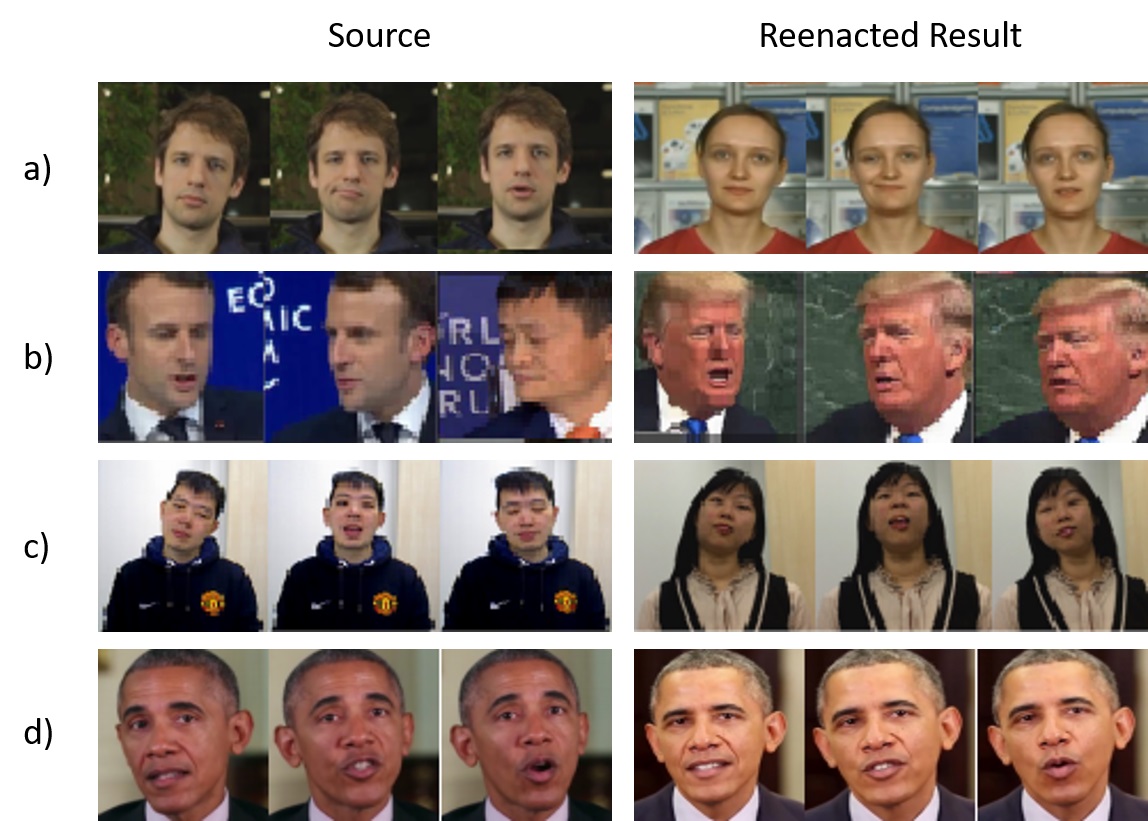}
\end{center}
   \caption{Effect of reenactment on the target sequence by (a) Kim \etal \cite{kim2018deep}, (b) Wu \etal \cite{wu2018reenactgan}, (c) Thies \etal \cite{thies2018headon}, and (d) Suwajanakorn \etal \cite{suwajanakorn2017synthesizing}.}
\label{fig:renact_c}
\end{figure}

\subsection{Generation Techniques}\label{generation_tech}
For the past decade, there has been significant work on transforming target video either from the input audio or video). These have been aimed at different applications ranging from expression transfer from one video footage to another \cite{thies2015real,thies2016face2face}, lip-syncing of the target from input audio \cite{garrido2015vdub,suwajanakorn2017synthesizing}, and mimicking the movement of the source to target \cite{thies2018headon}. The effect of these works can be considered as reenactment manipulations, as the resultant movement of the target is modified in the process or has been \textit{reenacted} upon with. Figure \ref{fig:renact_c} depicts the effect of generation techniques upon the target actors.

Suwajanakorn \etal  \cite{suwajanakorn2017synthesizing} proposed an approach towards the generation of a photo-realistic video from a target video of President Obama and lip-syncing to the input audio. The authors suggest a simplistic Recurrent Neural Network-based approach to synthesize the mouth shape of the targeting the input audio. Synthesis is primarily performed on lower face regions including mouth, cheek, chin, and nose. Garrido \etal \cite{garrido2015vdub} have presented a system based upon the capture of the 3D face model of both dubbing and target actors and then using audio analysis on the dubbing actors for creating a photo-realistic 3D mouth model to be applied upon the target actors.
\begin{figure*}[htbp]
\begin{center}
\includegraphics[width=17cm]{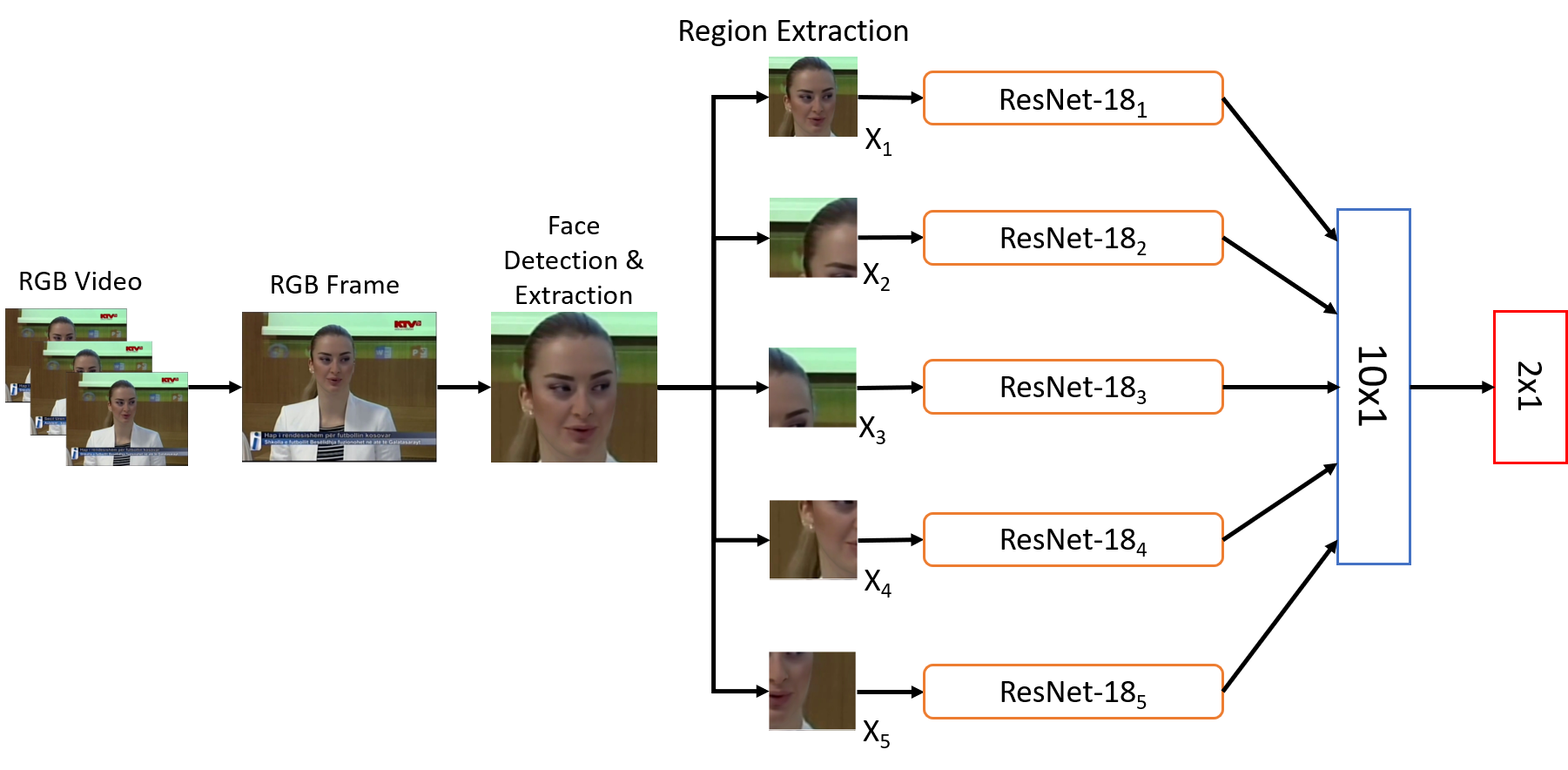}
\end{center}
   \caption{Proposed pipeline, RGB frames are sampled from RGB videos. ROI extraction is done on frames by face detection followed by local region extraction which acts as an input to the proposed classification network.}
\label{fig:pipeline}
\end{figure*}

Thies \etal\cite{thies2015real} have presented a method for real-time transfer of expression from one actor to another in a target sequence. Using RGB-D data as input, the proposed method keeps the non-face region unchanged while transferring the expressions. The authors presented a novel approach to represent facial identity and expression in a linear parametric model. The expressions are synthesized by changing the blend shape parameters of the target frames by the source. Thies \etal  \cite{thies2016face2face}, eliminated the need for depth videos in \cite{thies2015real}, thus allowing the transfer of expression to generic RGB videos (e.g. YouTube videos). Kim \etal  \cite{kim2018deep} presented a novel method of allowing full reanimation of portrait videos by the actor, including head pose, facial expression, eye motion, and in some cases, even the identity. The method employs the use of a face reconstruction approach to get a parametric representation of the face and illumination of each video frame. This representation is fed into a Render-to-video network based on the Conditional Generative Adversarial Network to generate the output frames. Wu \etal \cite{wu2018reenactgan} proposed reenactment through the transfer of facial features to a boundary latent space and then adapting the target boundary according to the source with the use of a transformer.
 Thies \etal \cite{thies2018headon} extended the concept reenactment to transfer of movement of the torso and head to the target video with the use of parametric models of the head, eye, and torso. These are later used to project the captured motion from the source to target in a photo-realistic fashion.

\subsection{Detection Algorithms}\label{detection_tech}
Work towards DeepFake detection has been sparse, due to the relatively new nature of the manipulation. However, the sheer degree of realism in the videos created by reenactment should have attracted more detection work in the field.

Afchar \etal \cite{afchar2018mesonet} proposed two shallow architectures in an attempt to capture the mesoscopic properties of images or frames. The first architecture Meso-4 comprises of four layers of convolution and pooling followed by a single-layered dense network. The other architecture MesoInception-4 performed modification on Meso-4 by replacing the first two convolution layers by a modified inception module. The authors also explored image aggregation on the proposed network in an attempt to better classify videos. Agarwal \etal \cite{agarwal2019protecting} suggested learning the head and facial movement of specific people of interest and then differentiating the movement in the DeepFake video of the same individual.

Face tampering detection techniques have been observed to be useful in the detection of manipulations by Face2Face. Zhou \etal \cite{zhou2017two} introduced a two-stream network, with one stream based on patch triplet stream with 5514D steganalysis features and other upon GoogleNet followed by score fusion of the two streams. Raghavendra \etal\cite{raghavendra2017transferable} have used feature level fusion by extracting features from fine-tuned VGG19 and AlexNet and are concatenated as input to Probabilistic Collaborative Representation Classifier. Bayer \etal \cite{bayar2016deep} have proposed a generic tampering detection algorithm, which is a shallow network of eight layers with a constrained CNN to suppress image content and adaptive learning of manipulation features. XceptionNet \cite{chollet2017xception}, which is based upon depth-wise separable convolution layers has also been shown to perform well for the detection task \cite{rossler2018faceforensics}.

\section{Proposed Detection Algorithm}
\label{method}
In this research, we have proposed a deep learning-based architecture for detecting reenacted frames generated using the Face2Face reenactment technique \cite{thies2016face2face}. The proposed method uses RGB frames in conjunction with a multi-stream network for improved extraction of localized facial artifacts and noise patterns introduced by the reenactment procedure. We also propose a loss function to facilitate balanced training of the proposed multi-stream network. The network captures local facial artifacts by the use of dedicated streams that learn their respective regional artifacts. A full-face stream then determines the dependency between the regions. By combined learning of regional and full-face artifacts, the proposed network can classify highly compressed frames with a relatively small drop in the performance as compared to the existing methods.

%  \begin{figure*}[htbp]
% \begin{center}
% \includegraphics[width=12cm]{Images/arch.png}
% \end{center}
%   \caption{Proposed Multi-stream Network Architecture.}
% \label{fig:arch}
% \end{figure*}
\subsection{Preprocessing}
 Figure \ref{fig:pipeline} shows the schematic representation of the proposed pipeline. The frames are extracted from the RGB video as per the experimental protocol defined in Section \ref{dataset}; this is followed by face detection by the S3FD approach \cite{zhang2017s3fd}. In the case where multiple faces are detected in a single frame, face mask annotations provided by the dataset are used to identify the target face. Mask annotations are used in case S3FD fails to detect the faces in the frame. This can be seen as a region of interest extraction step, which is also streamlined by strict square cropping centered around the face to suppress the background information as much as possible. For each face, the local region is extracted by dividing the frame into a $2\times2$ grid. This segregates the fundamental facial features into four regions, which is then followed by re-sizing each of the four local images and the full-face to $224\times224$.

\subsection{Network Architecture}
 As shown in Figure \ref{fig:pipeline}, the proposed multi-stream network consists of five parallel ResNet-18 models \cite{he2016deep} - four are dedicated to learning the local, regional artifacts and one for the overall effect of the reenactment upon the face. For each of the ResNet-18 models, the classification layer has been mapped to two outputs by a fully connected layer. The outputs from these five parallel ResNet-18 are concatenated to form a 10-dimensional vector which is passed upon to learn the weighted fusion of the scores for the binary classification task. 
 
The fundamental intuition is to make the network learn those features or artifacts that get suppressed when learning the model with only the full-face image. Training a model explicitly on a specific region of the image forces the network to learn those low-level spatial features that are not learned by the initial model, trained upon the full-face as shown in Figure \ref{fig:cam}, and can be used to improve the performance for the classification task specifically for highly compressed frames. It has been done keeping in mind the practicality of the problem. Since most of the time, manipulated videos or images that are circulated are in a highly compressed format, the drop in performance due to compression should ideally be as low as possible. Also, the prior knowledge that Face2Face manipulations affect the whole facial region adds to the improvement of the performance of the proposed network, as discussed in Section \ref{results}. A combination of such four regional models with the model trained on full image paves the way for a setup similar to Spatial Pyramid \cite{lazebnik2006beyond} structure. The two-level spatial pyramid has been taken, keeping in mind the need to maintain the balance between the model complexity and the information gain by the spatial features extracted by the model. The final fully connected layer learns the weighted mapping of scores of the four models trained upon the local regions and one model trained upon the full image.

\subsection{Loss Function}
 Let the input image be represented as $X$, and the corresponding output be $Y$ for the binary classification task i.e., classifying if the input $X$ has been manipulated or not. Each input $X = \{X_1,X_2,X_3,X_4,X_5\}$ is a set of five images of size $224\times224$ where $X_1$ represents the cropped full facial image and $X_2,X_3,X_4,X_5$ represent the four locally extracted images for each frame and $Y$ a binary value with $0$ denoting \textit{original} and $1$ denoting \textit{altered} frames. The following loss function is minimized during the training process.  
  
 \begin{figure*}[htbp]
\begin{center}
\includegraphics[width=13cm]{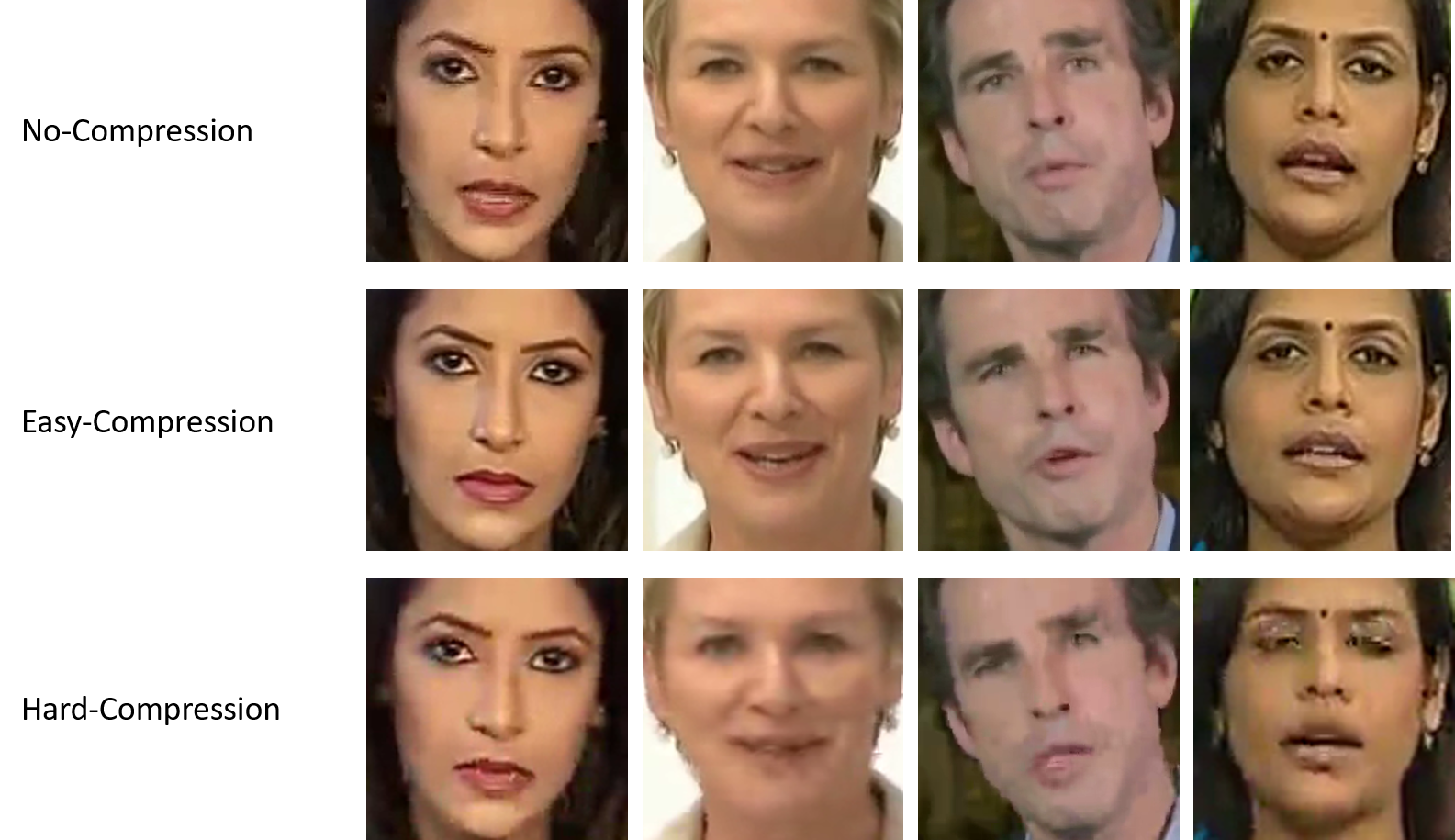}
\end{center}
   \caption{Illustrating the effect of compression on video frames.}
\label{fig:compression}
\end{figure*}

 \begin{equation}
     L_{total} = \underbrace{L_{R_1}}_\text{Full Face Loss} + \overbrace{\sum_{n=2}^{5} L_{R_i}}^\text{Local Regional Loss} +  \underbrace{ \lambda\times L_{fusion}}_\text{Fusion Loss}
 \end{equation}
 where $L_{total}$ is the effective loss and $L_{R_i}$ represents the cross-entropy loss as per Equation 2.
 
  \begin{equation}
 \label{cross_entro}
     L_{R_i} = -\sum_{c=0}^{1} Y_c\log f_c(X_i)
 \end{equation}
between the scores $f(X_i)$ of $ResNet_i$ model and true output $Y_c$. $L_{fusion}$ denotes the cross-entropy loss between the output of the final linear layer of the proposed model and the true output $Y$. The weight of $L_{fusion}$ is parameterized by $\lambda$. It is to be noted that during the calculation of various cross-entropy losses, the output of each model is first normalized using softmax in the range [$0,1$].

 The loss function has been designed to avoid the model from getting biased towards a particular ResNet model. Back-propagation of just $L_{fusion}$ as $L_{total}$ was found to be making the network biased towards $ResNet_1$ model, thereby reducing the performance of the network. By incorporating the loss of each of the parallel ResNet into the loss function, we prevent the network from being biased towards $ResNet_1$. Consequently, it improves the performance of the overall multi-stream network.
 
\subsection{Implementation Details}
The proposed network has been implemented with Python3.5 Pytorch deep learning framework. Optimization is performed using ADAM optimizer with default parameters ($\beta_1 = 0.9$ \& $\beta_2 = 0.999$) with a batch size of 32. The initial learning rate is kept at $10^{-4}$ and is divided by 10 after every 10 epochs. The ResNet-18 models are pre-trained on the ImageNet dataset \cite{deng2009imagenet} and then retrained on the face reenactment dataset. The value of the loss parameter $\lambda$ as 1 yields the optimal results.

\section{Dataset}
\label{dataset}

In this research, we have proposed a novel algorithm for detecting alterations that occur due to reenactment in RGB frames. For testing the performance of the proposed algorithm, we have used the FaceForensics Source-to-Target reenactment dataset \cite{rossler2018faceforensics}. The dataset is the only publicly available reenactment dataset for this task (FaceForensics++ \cite{rossler2019faceforensics++} also contains the same videos as in FaceForensics for reenactment detection task). The dataset consists of 1004 unique videos from YouTube. Each video sequence is at least 300 frames long at 30 fps. The videos have been modified using the Face2Face approach \cite{thies2016face2face} to produce reenactment manipulations. Therefore, for each video, the dataset contains the original video, reenacted video, and face mask against which the reenactment has been done. The dataset has been divided into train, test, and validation split as per Table \ref{tab:Dataset_Video}. For training and testing, we have followed the protocols mentioned in \cite{rossler2018faceforensics}, where 10 frames have been randomly sampled from each video, i.e., from 1004 original and altered videos. Thus, for each unique video, 20 frames have been sampled, 10 from original, and 10 from altered.

\begin{table}[tbp]
    \centering
     \caption{FaceForensics Dataset Composition}
\vspace{6pt}
    \begin{tabular}{|c||c|}
    \hline
    \textbf{Set}&\textbf{Number of Videos}\\
    \hline
    \hline
    \textbf{Train}& 704\\
    \hline
    \textbf{Validation} & 150\\
    \hline
    \textbf{Test} & 150\\
    \hline
    \end{tabular}
   
    \label{tab:Dataset_Video}
\end{table}

%The composition of the resultant sampled dataset is as per table. \ref{tab:Dataset_Frame}.
% \begin{table}[htbp]
%     \centering
%      \caption{Cardinality and composition of each split of  sampled dataset (frames)}
%     \begin{tabular}{|c|c|c|c|}
%     \hline
%     \textbf{Set}&\textbf{Original Class}&\textbf{Forged Class}&\textbf{Total}\\
%     \hline
%     \textbf{Train}& 7040 & 7040 & 14080\\
%     \hline
%     \textbf{Validation}& 1500 & 1500 & 3000\\
%     \hline
%     \textbf{Test}& 1500 & 1500 & 3000\\
%     \hline
%     \end{tabular}
   
%     \label{tab:Dataset_Frame}
% \end{table}

All the experiments have been performed on the dataset under three H.264 compression schemes with quantization parameter 0 for no-compression (no-c), 23 for easy-compression (easy-c), and 40 for hard-compression (hard-c). Compression has been performed to imitate the effect of compression of videos on various social media platforms such as Facebook and WhatsApp. The effects of compression can be seen in Figure \ref{fig:compression}.

\section{Results and Observations}
\label{results}
The proposed algorithm has been compared and contrasted with the respective state-of-the-art counterparts for the given dataset across various compression schemes. We have also analyzed multiple components of the proposed algorithm and its effect on the detection performance. The results are compared against the shallow network architecture such as MesoNet \cite{afchar2018mesonet} and Bayer \etal \cite{bayar2016deep}, state-of-the-art transfer learning architecture XceptionNet \cite{chollet2017xception} and face tampering detection algorithms like Zhou \etal \cite{zhou2017two} and Raghvendra \etal \cite{raghavendra2017transferable}. Baseline performance reported in \cite{rossler2018faceforensics} has been directly inferred for comparison. 

Table \ref{tab:result_comp} summarizes the accuracy of various reenactment detection algorithms on the FaceForensics dataset across the three compression modes. The classification accuracy has been calculated as the average of class-wise classification accuracies. The proposed model yields the best classification performance on the test set, it has a mean classification accuracy of 90.40\% with a standard deviation of 0.30\%. The results and analysis are discussed below.

\begin{itemize}
     \item With an increase in the degree of compression, there is a significant drop in the performance of all of the methods, as shown in Table \ref{tab:result_comp}. However, the decline in performance is high for shallow networks, such as MesoNet \cite{afchar2018mesonet} with four convolutional layers and a classification layer, and universal manipulation algorithm like \cite{bayar2016deep} with eight convolutional layers. In contrast, the drop in performance is comparatively lower for deep networks such as XceptionNet \cite{chollet2017xception} and two-stream network \cite{zhou2017two} with GoogleNet classification stream and steganalysis features as the second stream.
   
    \item Most of the detection methods give high performance on images with no compression or easy compression. However, the performance significantly reduces in the case of images with hard compression. This may have been caused because no-compression Face2Face manipulation tends to show edges around the corners of the chin and near the nostrils. Thus, allowing networks to quickly learn the difference between the original and the altered images. However, with compression, these details tend to vanish, and it becomes more and more challenging to learn the difference between the two classes. Figure \ref{fig:roc} also showcases the effect of compression upon the proposed network.

    \begin{table}[tp]
    \centering
    \caption{Accuracy (\%) of different algorithms on the FaceForensics dataset with different compression factors.}
 \vspace{6pt}
   \begin{tabular}{|l|c|c|c|}
   
        \hline
        \textbf{Methods} & \textbf{no-c} & \textbf{easy-c} & \textbf{hard-c}\\
        \hline 
        MesoNet, Afchar \etal \cite{afchar2018mesonet} & 96.80 & 93.40 & 83.20\\
        \hline
        Bayer \etal \cite{bayar2016deep} & 99.53 & 86.10 & 73.63\\
        \hline
        Zhou \etal \cite{zhou2017two} & 99.93 & 96.00 & 86.83\\
        \hline
        Raghvendra \etal \cite{raghavendra2017transferable} & 97.70 & 93.50 & 82.13\\
        \hline
        XceptionNet \cite{chollet2017xception} & 99.93 & 98.13 & 87.81\\
        \hline
        Proposed Approach & \textbf{99.96} & \textbf{99.10}  & \textbf{91.20}\\
        \hline
        
    \end{tabular}
   
    \label{tab:result_comp}
\end{table}

\begin{table}[t]
    \centering
    \caption{Classification performance (\%) of ResNet and VGG models on the FaceForensics dataset.}
  \vspace{6pt}
  \begin{tabular}{|l|c|c|c|}
         \hline
        \textbf{Network} & \textbf{no-c} & \textbf{easy-c} & \textbf{hard-c}\\
        \hline 
        VGG16 \cite{simonyan2014very} & 99.50 & 96.90 & 85.20\\
        \hline
        ResNet-18 \cite{he2016deep} & 99.93 & 97.70 & 88.20\\
        \hline
        ResNet-50 \cite{he2016deep} & 99.93 & 97.40& 86.40\\
        \hline
        ResNet-152 \cite{he2016deep} & 99.89 & 97.60 & 85.70\\
        \hline
        Proposed Approach & \textbf{99.96} & \textbf{99.10}  & \textbf{91.20}\\
        \hline
    \end{tabular}
    
    \label{tab:res_pretrained}
\end{table}
    \item As can be seen from Table \ref{tab:res_pretrained}, increasing the layers do not specifically improve the classification performance. The models give consistent, comparable performance in case of no or easy compression, but a significant drop in performance is observed in the case of a network with a high number of layers with frames compressed with high quantization factors. This may be due to the inability of ResNet-50 and ResNet-152 to learn its large number of model parameters optimally as compared to ResNet-18 when there is a significant loss of information in the input, which is the resultant effect of severe compression.
    
\begin{figure}[tp]
\begin{center}
\includegraphics[width=8cm]{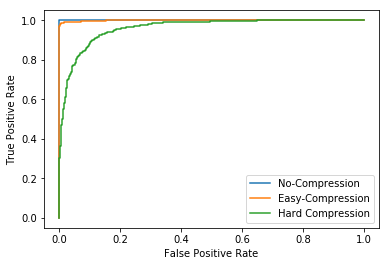}
\end{center}
   \caption{ROC Curves of the proposed network for different compression modes.}
\label{fig:roc}
\end{figure}

     \begin{figure*}[tbp]
        \begin{center}
        \includegraphics[width=13cm]{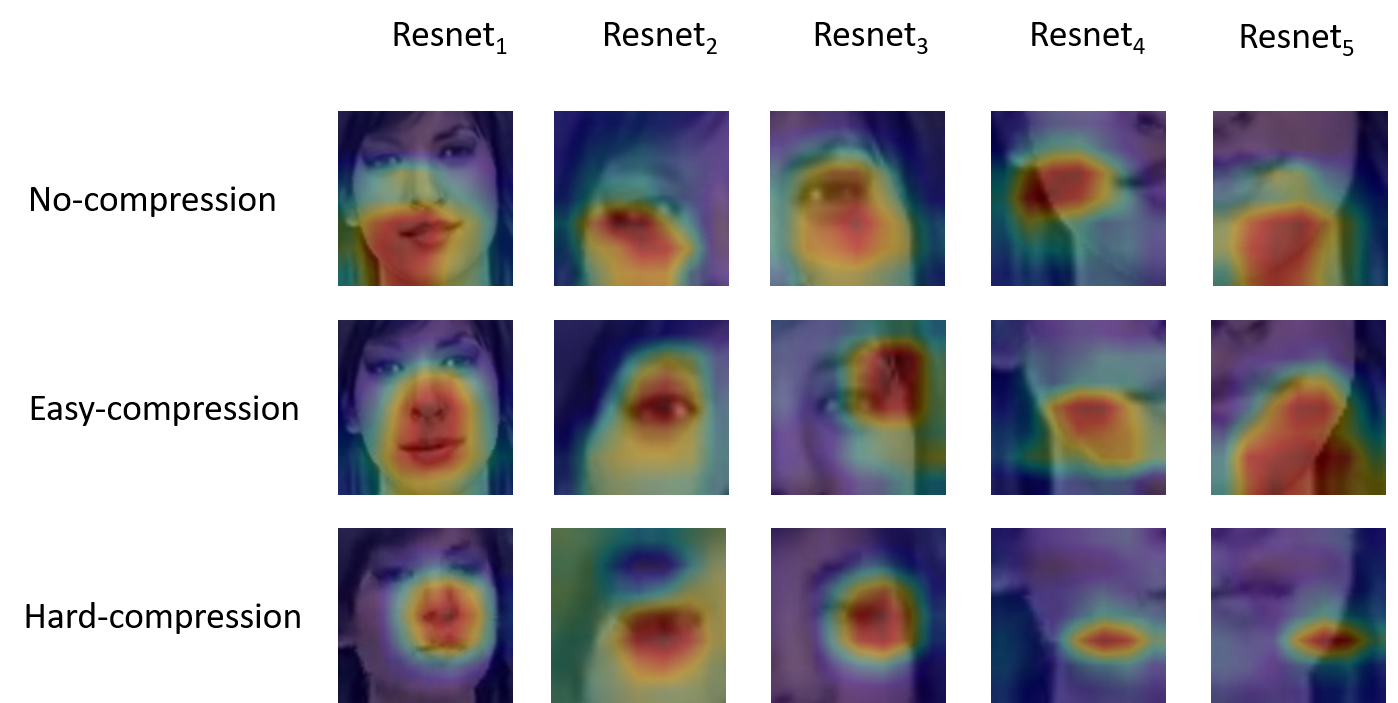}
        \end{center}
        \caption{Class activation maps for local and full face ResNet for the Proposed Network.}
        \label{fig:cam}
    \end{figure*}   
    
 \begin{table}[tp]
        \centering
        \caption{Score fusion results (\%) for hard-c compression.}
\vspace{6pt}
        \begin{tabular}{|l|l|c|}
             \hline
        \textbf{Classifiers} & \textbf{Fusion} & \textbf{Accuracy}\\
        \hline 
      Regional Classifiers & SVM & 85.80\\
        \hline
        Regional Classifiers & Proposed & \textbf{88.26}\\
        \hline
        All Classifiers & SVM & 89.13\\
        \hline
        All Classifiers & Neural Net & 89.80\\
        \hline
        All Classifiers & Proposed & \textbf{91.20}\\
        \hline
        
        \end{tabular}
        
        \label{tab:res_fusion}
    \end{table}
\item Table \ref{tab:res_fusion} showcases the performance of streams under various fusion techniques and also the effectiveness of the proposed loss function. It is observed that the fusion of output scores of independently trained region-based ResNet models by Linear-SVM, performs comparably to ResNet-50 and ResNet-152. Thus, presenting quantitative evidence of discriminative features available in these regions. The fusion of scores of all ResNet models gives a significant boost to the classification performance. It is also to be noted that a linear combination of scores emulated by a single-layered neural network slightly outperforms the support vector machine based score fusion. End-to-end training of the proposed architecture with binary cross-entropy function gives a classification score similar to the score fusion by a neural network. However, end-to-end training by the proposed loss function further improves the performance of both networks, i.e., with only regional classifiers and the proposed architecture.

\begin{table}[t]
    \centering
    \caption{Classification performance (\%) of the proposed network on cross-compression. The rows and columns depict the compression mode of train and test set respectively.}
\vspace{6pt}
    \begin{tabular}{|c|c|c|c|}
         \hline
        \multirow{2}{*}{\textbf{Network Trained On}} & \multicolumn{3}{c|}{\textbf{Network Tested On}} \\ \cline{2-4}
            & \textbf{no-c} & \textbf{easy-c} & \textbf{hard-c} \\\hline 
        No-Compression & 99.96 & 58.26 & 52.66\\
        \hline
        Easy-Compression & 99.56 & 99.10 & 55.43\\
        \hline
       Hard-Compression & 96.73 & 95.76 & 91.20\\
        \hline
    \end{tabular}
    \label{tab:cross_compression}
\end{table}

\item Table \ref{tab:cross_compression} summarizes the performance of the proposed architecture on cross-compression dataset. The network is thereby trained upon frames of one compression mode and then tested against frames compressed by different compression modes. It is observed that the performance of models trained upon low compression drops significantly for the input of higher compression. However, the models trained upon highly compressed frames generalize better across the low compressed inputs.
    % \newpage
    \item  We analyze the class activation maps, as shown in Figure \ref{fig:cam} for the regional and full-face classifiers across the compression schemes.

    \begin{itemize}
       \item Class activation maps corresponding to full-face trained ResNet, i.e., $ResNet_{1}$ indicate that the nose and mouth regions provide the fundamental differentiation between the original and altered frames. This is because during the process of reenactment, the realistic portrait of movement of mouth and nearby regions are hardest to create as the transfer of static features are easy to perform as compared to dynamic features. Also, the lower facial regions are more prone to movement as compared to any other facial regions in a video sequence.
 
        \item The drop in performance of the proposed multistream network with respect to compression factors can easily be inferred from the activation maps. The higher the compression factor, the smaller is the activated region of the network trained for the classification task.
        
        \item Unlike the forehead regions, which are not prone to high movement, the full-face ResNet fails to detect the artifacts generated due to the movement in the eye region. This shows the need of local classifiers dedicated for alteration detection near the eye region by $ResNet_2$ and $ResNet_3$.

        \item Face2Face approach incorporates blendshape detection followed by parametric transfer of facial expression, thus leading to the creation of edge artifacts near the face boundary due to the error in face tracking and effective transfer of expression. These errors are again neglected by the full-face ResNet specifically in case if compression is applied upon the input video. $ResNet_4$ and $ResNet_5$, thus help in providing another indicative measure of falsification by exploiting the face tracking limitations of the Face2Face approach.    
    \end{itemize}
    
    \begin{table}
        \centering
         \caption{Classification accuracy (\%) of regional classifiers for frames with hard-compression.}
        \begin{tabularx}{\linewidth}{|c|c|c|}
        \cline{1-3}
        \textbf{} &\textbf{Stream}& \textbf{Accuracy}\\
        \cline{1-3}
        \multirow{5}{*}{Individual} &Face (X1) & 88.20\\
        \cline{2-3}
                                    & Left Eye (X2) & 78.95\\
        \cline{2-3}
                                    & Left Cheek (X4) & 79.15\\
        \cline{2-3}
                                    & Right Eye (X3) & 77.45\\
        \cline{2-3}
                                    & Right Cheek (X5) & 74.20\\
        \cline{1-3} 
        
        \multirow{5}{*}{Combination}& Regional & 88.26\\
        \cline{2-3}
            & Face + Left Eye & 88.60\\
        \cline{2-3}
        &Face + Left Cheek & 89.30\\
        \cline{2-3}
        &Face + Right Eye & 88.83\\
        \cline{2-3}
        &Face + Right Cheek & 88.40\\
        \cline{1-3}
        \end{tabularx}
       
        \label{tab:patch_wise}
    \end{table}
    
    \item Table \ref{tab:patch_wise} summarizes the classification accuracy of the classifiers for the individual as well as the combination of streams in the proposed model. It is observed that the left-sided features perform better than the right-sided features, specifically in the case of the cheek region. It is also observed that the eye region gives a more consistent classification performance than the cheek region. This can be inferred from the class activation maps as the regional classifiers are able to extract more prominent features in the eye region than the cheek region across all the compression schemes. The regional classifiers combined have performance comparable to the full-face classifier. Also, the contribution of each stream in combination with the full-face classifier is proportional to the classification performance of each regional classifier.
    
    \item We analyzed the effect of parameter $\lambda$ upon the classification performance. The proposed network yields a classification accuracy of $89.00\%$, $91.20\%$ and $87.50\%$ for $\lambda$ $= 0.001$, $1$, and $100$, respectively. A very small value of $\lambda$ is equivalent to training the streams independently and then fusing the scores whereas a high value of $\lambda$ depicts an end to end training with standalone cross-entropy loss at the output layer of the network. The best classification performance was achieved by $\lambda = 1$, i.e., the weight of fusion was equal to the weight of individual streams.
    
    \item Figures \ref{fig:o_alter} and \ref{fig:alter_o} show some instances where the proposed network is not able to correctly classify the majority of the frames of the input subjected to hard compression.

\end{itemize}

\begin{figure}[tbp]
\begin{center}
\includegraphics[width=8cm]{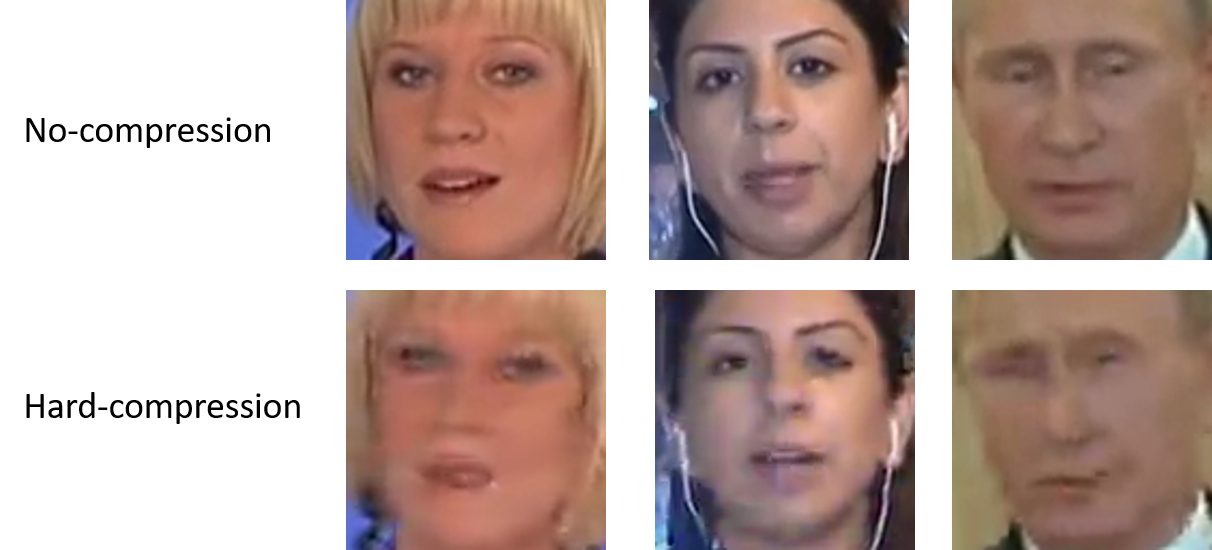}
\end{center}
  \caption{Frames misclassified as \textit{Altered} due to compression.}
\label{fig:o_alter}
\end{figure}
\begin{figure}[htbp]
\begin{center}
\includegraphics[width=8cm]{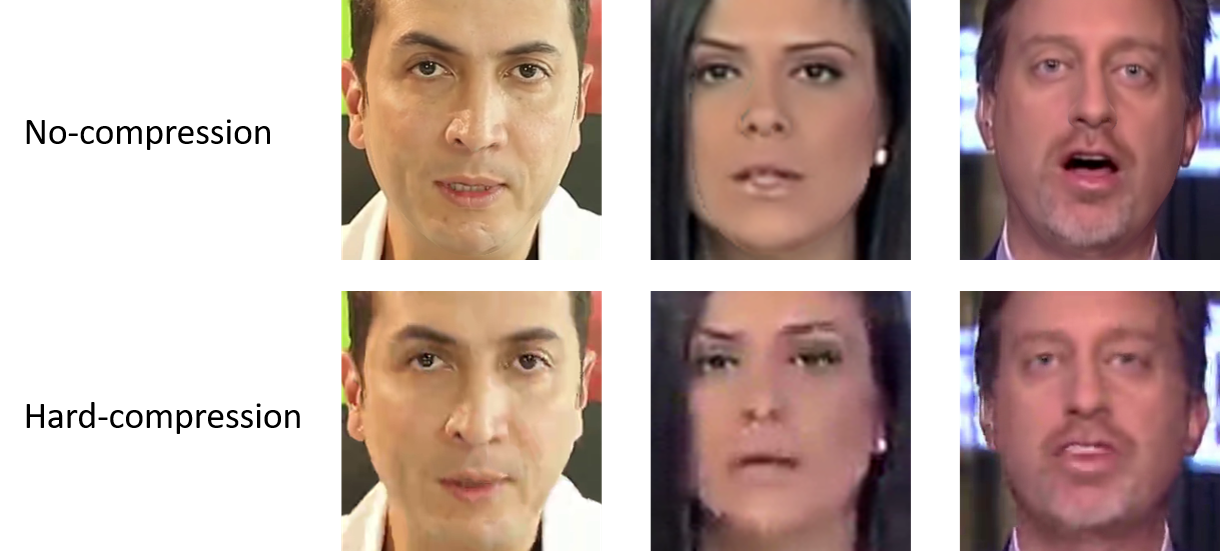}
\end{center}
  \caption{Frames misclassified as \textit{Original} due to compression.}
\label{fig:alter_o}
\end{figure}

\section{Conclusion and Future Work}
In this research, we have addressed reenactment based DeepFake detection in videos. The proposed detection algorithm outperforms state-of-the-art methods on the FaceForensics dataset and shows the smallest reduction in classification performance when the input video frame is subjected to adverse compression. In the proposed algorithm, we aim to find local noise patterns and artifacts that are left behind when altered with Face2Face reenactment. This allows the network to model itself upon various noise patterns learned by various regional classifiers, further aided by the full-face classifier. We also propose an end to end training loss function to allow for balanced training of regional classifiers as compared to the full-face classifier. Such type of loss function can find use in cases where the fusion of classifiers with different rates of convergence is needed. In order to develop a generalized detection approach, it is important to understand the DeepFake generation mechanism and try to leverage the limitations of various modules used in the generation of reenacted videos. The proposed model contains five parallel streams, thus leading to high computational complexity. In the future, we plan to reduce the model complexity by using an attention mechanism that learns the dependency between the image regions and features maps in a more computationally effective manner.

\section{Acknowledgement}
M.Vatsa is supported through the Swarnajayanti Fellowship by the Government of India. R. Singh and M. Vatsa are partly supported by the Ministry of Electronics and Information Technology, Government of India.

{\small
\bibliographystyle{ieee}
\bibliography{egbib}
}

\end{document}